\documentclass[
]{ceurart}

\usepackage{booktabs}
\usepackage{multirow}
\usepackage{graphicx}
\usepackage{caption}
\usepackage{bbding}

\usepackage{booktabs}
\begin{document}

\copyrightyear{2021}
\copyrightclause{Copyright for this paper by its authors.
  Use permitted under Creative Commons License Attribution 4.0
  International (CC BY 4.0).}

\conference{De-Factify: Workshop on Multimodal Fact Checking and Hate Speech Detection, co-located with AAAI 2022. 2022 
Vancouver, Canada}

\title{LAHM : Large Annotated Dataset for Multi-Domain and Multilingual Hate Speech Identification}

\author{Ankit Yadav}[%
email=ankit.yadav@logically.co.uk,
]
\author{Shubham Chandel}[%
email={Shubham.c@logically.co.uk},
]
\author{Sushant Chatufale}[%
email=sushant.c@logically.co.uk,
]
\author{Anil Bandhakavi}[%
email=anil@logically.co.uk,
url=https://www.logically.ai/team/leadership/anil-bandhakavi
]
\address{Logically.ai, Brookfoot Mills, Brookfoot Industrial Estate, Brighouse, HD6 2RW, United Kingdom}

\begin{abstract}
Current research on hate speech analysis is typically oriented towards monolingual and single classification tasks. In this paper, we present a new multilingual hate speech analysis dataset for English, Hindi, Arabic, French, German and Spanish languages for multiple domains across hate speech - Abuse, Racism, Sexism, Religious Hate and Extremism. To the best of our knowledge, this paper is the first to address the problem of identifying various types of hate speech in these five wide domains in these six languages. In this work, we describe how we created the dataset, created annotations at high level and low level for different domains and how we use it to test the current state-of-the-art multilingual and multitask learning approaches. We evaluate our dataset in various monolingual, cross-lingual and machine translation classification settings and compare it against open source English datasets that we aggregated and merged for this task. Then we discuss how this approach can be used to create large scale hate-speech datasets and how to leverage our annotations in order to improve hate speech detection and classification in general.
\end{abstract}

\maketitle


\section{Introduction}

Abusive language is an important and relevant issue in social media platforms such as Twitter. Social media is often exploited to propagate toxic content such as hate speech or other forms of abusive language. The amount of user-generated content produced every minute is very large, and manually monitoring abusive behavior in Twitter is not feasible and impractical. Twitter has made efforts to eliminate abusive content from their platform by providing clear policies on hateful conduct, user reporting  and using moderators to filter content. Still, these manual efforts are not scalable enough and are not long term.

Several studies from the Natural Language Processing (NLP) field have been done to tackle the problem of hate speech detection in social media. Most studies proposed a supervised approach to detect abusive content automatically using various models ranging from traditional machine learning approaches to deep learning based approaches. However, the majority of work focused only on a single language, i.e., English, and a single abusive domain phenomenon, e.g., hate speech, sexism, racism, religious hate and so on, rather than multiple languages and multiple domains.
Twitter supports content in 34 languages and user can use any one of them to express views. Thus the problem of tackling hateful content in real time in multiple languages becomes a challenge.
We need robust models for hateful content detection across multiple languages and multiple domains.\\
In this paper we try to tackle two prominent challenges in hate speech detection-
\begin{enumerate}
\item Build a \textbf {multilingual dataset} for hate speech detection across 6 languages -: \emph{English, Hindi, French, Arabic, German, Spanish.}
\item Build a \textbf{multi-domain dataset} that covers these hate speech domains -: \emph{ Abusive, Racism, Sexism, Religious Hate and Extremism}
\end{enumerate}
We define the different domains as follows-:
\begin{enumerate}
\item RACISM: Discrimination based on race, ethnicity, caste, nationality, culture, skin colour, hair texture, physical aspects.
\item SEXISM: Discrimination based on gender/sexual orientation.
\item RELIGIOUS HATE: Religious discrimination treating a person or group differently because of the particular faith/belief which they hold about a religion.
\item ABUSE: Speech that causes or likely to cause distress, disrespect or mental pain, especially from vulgar and profane comments.
\item EXTREMISM: Speech that cause or is likely to cause, harm to individuals, communities or wider society, and where any political, civil issues can lead to extremist behaviour through violence.
\end{enumerate}

\section{Related Work}
There have been several studies on abusive language detection \cite{ousidhoum-etal-multilingual-hate-speech-2019} \cite{chung2019conan}, offensive language, hate speech identification\cite{davidson2017automated}, toxicity  \cite{kolhatkarcorpus}, hatefulness \cite{gao2017detecting}, aggression \cite{kumar2018benchmarking}, attack \cite{wulczyn2017ex}, racism, sexism \cite{waseem-hovy:2016:N16-2}, obscenity, threats, and insults.

Along with that, there are several shared tasks that have focused on abusive language and hate speech detection such as HASOC-2019 \cite{mandl2019hasoc}, TRAC shared task on aggression identification \cite{kumar2018benchmarking}, HatEval \cite{basile2018crotonemilano} and GermEval-2018 \cite{wiegand2018overview}which focused on offensive language identification in German tweets.

{Waseem}\cite{waseem-hovy:2016:N16-2} proposed the following list to identify hate
speech. Their criteria are partially derived by negating the privileges observed in McIntosh (2003),
where they occur as ways to highlight importance,
ensure an audience, and ensure safety for white
people, and partially derived from applying common sense.
A tweet is categorized as offensive if it:
\begin{enumerate}
\item uses a sexist or racial slur.
\item attacks a minority.
\item seeks to silence a minority.
\item criticizes a minority (without a well founded
argument).
\item promotes, but does not directly use, hate
speech or violent crime.
\item criticizes a minority and uses a straw man argument.
\item blatantly misrepresents truth or seeks to distort views on a minority with unfounded
claims.
\item shows support of problematic hash tags - “\#BanIslam”, “\#whoriental”, “\#whitegenocide”
\item negatively stereotypes a minority.
\item defends xenophobia or sexism.
\item contains a screen name that is offensive, as
per the previous criteria, the tweet is ambiguous (at best), and the tweet is on a topic that
satisfies any of the above criteria.
\end{enumerate}

Wide variety of machine learning models have been used to deal with multi-domain hate speech detection task. Some studies use traditional machine learning approaches such as, logistic regression \cite{salminen2020developing}, support vector machine \cite{chowdhury2020multi} \cite{wiegand2018inducing}, linear support vector machine classifiers (LSVC) \cite{karan2018cross} \cite{pamungkas2019cross} \cite{pamungkas2020misogyny}. They used it for better explainability, along with several deep learning based models, including convolutions neural networks \cite{meyer2019platform} \cite{wang2020detect}, LSTM \cite{arango2019hate} \cite{meyer2019platform} \cite{pamungkas2020misogyny} \cite{pamungkas2019cross} \cite{waseem2018bridging}, bidirectional LSTM \cite{rizoiu2019transfer}. 
Most recent works focus on transfer learning and novel architectures involving Transformers based models such as  Bidirectional Encoder Representations from Transformers (BERT) \cite{caselli2020hatebert} \cite{koufakou2020hurtbert} \cite{glavavs2020xhate} \cite{mozafari2020hate} \cite{ozler2020fine} and its variants like RoBERTa \cite{glavavs2020xhate} in the cross-domain abusive language detection task.

For cross-lingual abusive language detection, most studies utilized transformers based models. Some traditional models used such as logistic regression \cite{aluru2021deep} \cite{basile2018crotonemilano} \cite{vashistha2021online}, linear support vector machines \cite{pamungkas2019cross} \cite{pamungkas2020misogyny}, SVM \cite{ibrohim2019translated}, 
LSTM \cite{corazza-etal-2020-hybrid} \cite{vashistha2021online} and Bi-LSTMs \cite{corazza-etal-2020-hybrid} have also been used. 
Recent work focused on several transformer based architectures such as multilingual BERT  \cite{ahn-etal-2020-nlpdove} \cite{aluru2021deep} \cite{glavavs2020xhate} \cite{pamungkas2020misogyny} \cite{perez-etal-2020-andes} \cite{stappen2020cross} \cite{vashistha2021online}, RoBERTa \cite{dadu-pant-2020-team}, XLM \cite{corazza-etal-2020-hybrid} \cite{stappen2020cross} and XLM-RoBERTa \cite{dadu-pant-2020-team} \cite{glavavs2020xhate} \cite{ranasinghe-zampieri-2020-multilingual}.
Transformers based models with multilingual language representations can easily deal with language shift in zero-shot cross-lingual task. 

\section {LAHM Dataset}

In this section, we describe the characteristics that we want to include in our dataset, our approach to collect different types of hate speech while covering all major hate speech domains and how to annotate data at large scale. We also give detailed statistics and analysis for the collected data.

\begin{figure*}
  \centering
  \includegraphics[width=\linewidth]{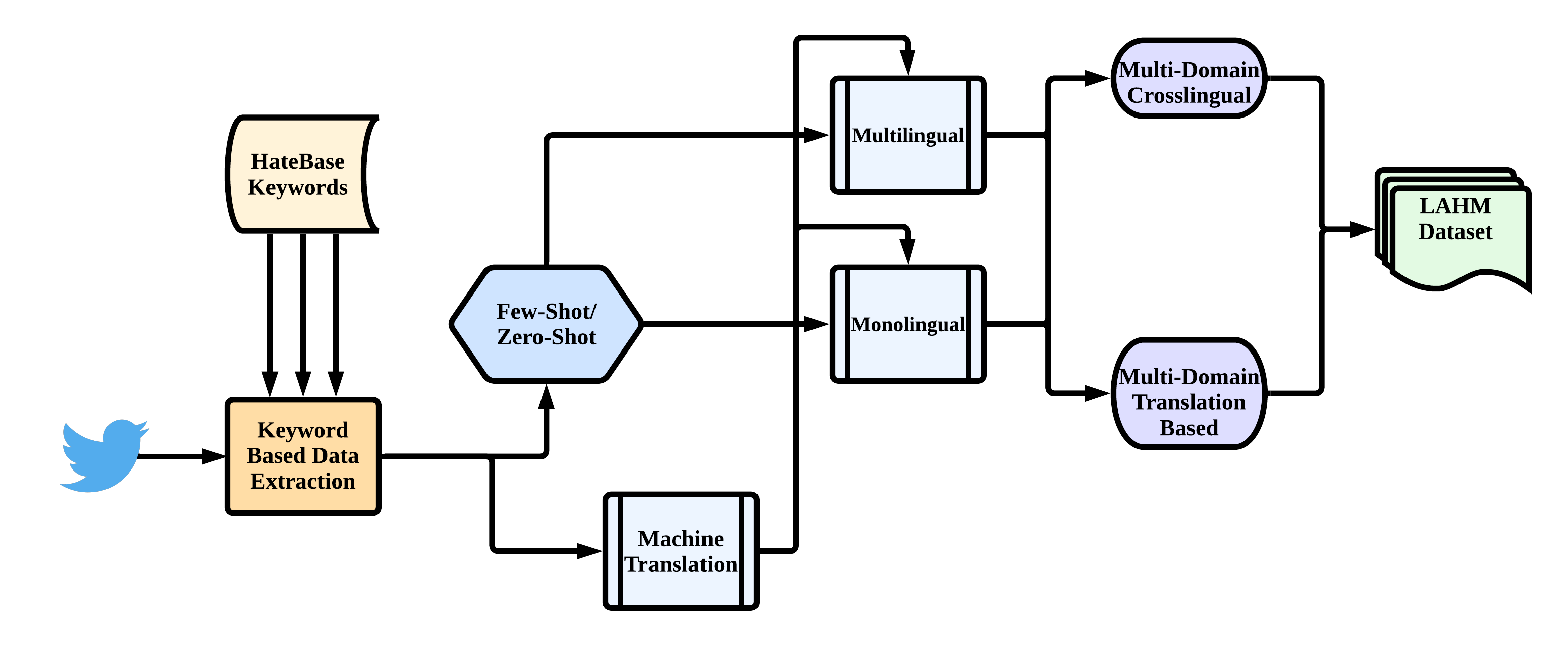}
  \caption{Overall Annotation Pipeline Architecture}
  \label{arch}
\end{figure*}

\subsection{Essentials of LAHM dataset}

Considering no such dataset is available at present that covers these five  domains of hate speech in the six languages, our aim is to create a reliable multilingual and multifaceted hate speech dataset.
\paragraph{Multilingual data:}Our dataset is created as multilingual resource to facilitate cross-lingual research. It contains hate speech in English, Hindi, French, Arabic, German and Spanish languages.

\paragraph{Multi-Domain data:}Our dataset consists of  fine-grained labels for each hate sample per language. These aspects cover majority of hate domains such as racism, sexism, abuse, religious hate and extremism.\\ 

\subsection{Dataset Collection}
\subsubsection{Language and Domain Specific Keywords Collection}

Considering the cultural differences in the main regions where English, Hindi, French, Arabic, Spanish and German are spoken, we start by looking for the hateful keywords that are native to these languages. 
We use \hyperlink{hatebase.com}{HateBase} vocabulary dataset, which is a valuable lexicon for creating hateful dataset from public forums, as well as Hatebase's sightings dataset, which is useful for trending analysis of keywords. 

We targeted 6 languages. Additionally we tried some indic languages mainly Marathi and Bangla but hatebase has negligible coverage for these languages in terms of keywords.
Additionally, we extracted the following:
\begin{enumerate}
\item Targeted groups for the keywords
\item Offensive levels of keywords - \emph{Extremely, Highly and Mildly offensive.}
\item  Recent sighting counts
\end{enumerate}

Target groups help us categorise keywords in low level classes of ethnicity \& nationality, religious hate, gender \& sexual orientation and general abuse.
Details of keywords per domain and language are provided in Table \ref{table 1} in dataset statistics. Total keywords collected were \textbf{1022}  . \\
To extract data for extremism, we used a set of keywords related to extremism and terrorism (including terrorist organisation names) to retrieve news articles from BBC Monitoring\footnote{\url{https://monitoring.bbc.co.uk/}}.
\begin{enumerate}
    \item  We further extracted comments from those articles related to extremism and terrorism. We also extracted tweets and comments which were extremist in nature from the counterextremism database\footnote{\url{https://www.counterextremism.com/daily-dose-archive}}.
    \item The comments extracted were said by members and suspects of terrorist and extremist organisations on Facebook, Twitter and YouTube.
\end{enumerate}

\subsubsection{Large Scale Multilingual Dataset of Tweets}
We started our dataset collection by using the keywords built per language. We utilize twint\footnote{\url{https://github.com/twintproject/twint/}} API to collect 1000-2000 tweets per keyword as Twitter official API can only be utilized for limited number of requests. We searched through the API for last one year of data. We also add additional keywords from  MLMA \cite{ousidhoum-etal-multilingual-hate-speech-2019} for Arabic and French.

The collected raw data contains cross-lingual tweets and therefore language detection becomes a part of our process. For each keyword in each language we consider tweets only in that language and drop the rest. This helps us in training monolingual hate speech detection without any need to worry for code switching in languages. 

We substituted all usernames with @USER and urls with @URL and cleaned any unnecessary symbols from the tweet. We also discarded short tweets with less than 4 words. In total we collected \textbf{497660} tweets. Details of tweets per language is given in section 3.3 \textit{Dataset} \textit{Statistics}.

For extremism class we carefully handcrafted a total of \textbf{88} keywords and collected the data from BBC monitoring. Additional data was collected from other websites for extremism. 

\subsection {Dataset Statistics}
Table \ref{table 1}  gives us the distribution of keywords across different languages and different domains. We have merged classes from hate base to arrive at those domains. \textit{Sexism} contains keywords that belong to \textit{gender} and \textit{sexual orientation}. Racism includes classes of \textit{ethnicity} and \textit{nationality}. Religious Hate includes words belonging to Religion category on Hatebase \footnote{\url{https://hatebase.org/}}. Extremism keywords were handcrafted as described in section 3.2.

English was the dominant language in terms of the total keywords (500) extracted. Besides Hindi, \textit{racism} is the major domain for which all languages have highest number of hate keywords.  

\begin{table*}[t]\centering
\caption{Keywords  language \& domain distribution}
\label{table 1}
\begin{tabular}{@{}lllllll@{}}\toprule
Language    &   Sexism & Racism & Abuse  & Religious Hate &  Extremism \\ \cmidrule(r){1-6}
English     &     69   &  418   & 30  &  33              &     88      \\
Hindi       &     75   &  16    &  9  &   8              &     -       \\
Arabic      &     27   &  35    &  17  &   17            &     -       \\
French      &     13   &  85    & 13   &  9              &     -       \\
German      &     21   &  67    & 13   &  11             &     -       \\
Spanish     &     43   &  77    & 21   &   2             &     -        \\\cmidrule(r){1-6}
\textbf{Total} & \textbf{248} & \textbf{698} & \textbf{103} & \textbf{80} & \textbf{88}\\
\bottomrule
\end{tabular}
\end{table*}

\begin{table*}[t]
\centering
\caption{Tweets distribution}
\label{tab2}
\begin{tabular}{@{}|l|l|l|l|@{}}
\toprule
\textbf{Languages} & \textbf{Raw} & \textbf{Processed} \\ \midrule
English            & 184339                     & 105120             \\ \midrule
Hindi              & 59321                       & 32734              \\ \midrule
Arabic             & 27374                       & 5394               \\ \midrule
French             & 75126                      & 20809              \\ \midrule
German             & 30868                        & 8631               \\ \midrule
Spanish            & 120632                        & 55148              \\ \midrule
\textbf{Total}     & \textbf{497660}                   & \textbf{227836}             \\ \bottomrule
\end{tabular}

\end{table*}
\subsection{HSMerge Dataset}
\subsubsection{Dataset Preparation} 
In order to build the dataset for different tasks with gold labels, we utilize 10 publicly available datasets for different types (tasks) of hate speech detection. We sampled annotated examples from diverse English
datasets: GAO hatefulness \cite{gao2017detecting}, TRAC aggression \cite{kumar2018benchmarking}, offensive \cite{zampieri2020semeval}, racism and sexism \cite{waseem-hovy:2016:N16-2}, MLMA \cite{ousidhoum-etal-multilingual-hate-speech-2019}, attack \cite{wulczyn2017ex}, CONAN \cite{founta2018large}, MMHS150k \cite{gomez2020exploring}.

We map labels from these datasets into different domains: abusive, non-abusive, sexism, racism and religious hate. For OLID \cite{zampieri2020semeval} \textit{OFFENSIVE} maps to \textit{Abuse}. GAO and WUL had binary labels, while the original TRAC uses three labels: non-aggressive, covertly-aggressive, and openly-aggressive. We relabel the first as non-abusive, and the other two as abusive. For MLMA we create labels on the basis of target groups: target group \textit{RELIGION} becomes \textit{RELIGIOUS HATE}. For MLMA \cite{ousidhoum-etal-multilingual-hate-speech-2019} target group \textit{GENDER} and \textit{SEXUAL ORIENTATION} maps to \textit{SEXISM}, target group \textit{RELIGION} becomes \textit{RELIGIOUS HATE}, target group \textit{ORIGIN} becomes \textit{RACISM}.
For HASOC \cite{mandl2019hasoc} \textit{OFFENSIVE} maps to \textit{ABUSE}. For COnan\cite{fanton-2021-human} target groups \textit{MUSLIMS}, \textit{JEWS} becomes \textit{RELIGIOUS HATE} labels, target groups \textit{MIGRANTS}, \textit{POC} maps to \textit{RACISM}, target group\textit{ WOMEN, LGBT+} maps to \textit{SEXISM} and \textit{DISABLED} maps to general \textit{ABUSE} group.
For MMHS150K \cite{gomez2020exploring} we got \textit{RACISM, SEXISM, RELIGIOUS HATE \& OTHERHATE}.

For some of these data we had to hydrate tweets from Twitter using official Twitter API as only the tweet ids had been provided. For Waseem and Founta \cite{founta2018large}, majority of the tweets were not available on Twitter which led to less number of tweets compared to the official dataset. We standardized these datasets into a single \textit{HSMerge} dataset. Details of classes and total samples is available in Table \ref{tab hsmerge}.

\begin{table*}[t]\centering
\caption{HSMerge  open source gold dataset}
\label{tab hsmerge}
\begin{tabular}{@{}lllll@{}}
\toprule
Datasets & Domain &  Hate & Non-Hate  &  Total             \\ \cmidrule(r){1-5}
OLID     &   Offensive / NotHate & 4400 & 8840 &    13240     \\

COnan  & Racism/ Sexism/ Religious/ Abuse/ OtherHate & 938/1025/1921/175/179 & - &  4238 \\
TRAC  & Hateful / NotHate &  5993 & 4348 & 10341\\
MLMA & Racism/ Sexism/ Religious/ OtherHate & 2448 / 1152 / 68/1979  & - & 5647  \\
MLA150K & Racism/ Sexism/ Religious/ OtherHate/ NotHate  &  11925 / 7365 / 163 / 5811    & 112845 &         138209   \\
HASOC & Hateful / NotHate  & 2261 & 3591 & 5852  \\
 WASEEM & Racist/Sexist/NotHate & 2753/15    & 7640 &10408  \\
WUL & Hate/NotHate & 8817 & 62937 & 71754  \\
GAO &  Hate / NotHate   & 244 & 675  & 909   \\

\bottomrule
\end{tabular}
\end{table*}

\section {Experiments}
\subsection{Overall Methodology and Models}
In this section, we describe the various components of our pipeline as shown in figure \ref{arch} and the models used for semi-supervised annotation process in different settings. Most of the hate speech detection tasks depend on manual annotation that limits the number of samples that can be labeled.
We used a hierarchical approach to validate and refine our initial keyword based dataset from Twitter.

\noindent
\begin{enumerate}
\item Monolingual hate speech detection models from Hate-ALERT used as zero shot pipeline to get high level labels of \emph{Hate} or \emph{NoHate} in monolingual settings.
\item Multilingual \textit{mbert} fine-tuned model to get high level labels.
\item Translate the raw data of five languages (other than English) to English.
\item Use multilingual binary classifier (developed in step 2) on translated data to classify tweets at high level.
\item Utilize Google Perspective API to predict toxicity of tweets for high level labels.
\item Use two way voting to predict final binary labels.
\item Fine-tuned "distilbert-base-uncased" model on only the English dataset, and used it to do annotations for the English translations of the data collected from Twitter in 5 languages.
\item To validate our domain specific labels we trained various models on HSMerge data shown in Table \ref{tab hsmerge} and predicted on hateful samples obtained from step 6.
\end{enumerate}
\subsection{Experimental Setting} 
To evaluate the performance of the models, we used weighted average F1 for benchmarking on validation set. All the experiments were done on NVIDIA A100 GPU with up to 20 GiB RAM.

\subsection{Hatefulness (Hate/No-Hate)}
We utilised hate speech models to perform initial validation on the raw LAHM dataset; details can be found in Section 5.

\subsubsection{Monolingual Experiments}
We utilized BERT based language specific hate speech models from HuggingFace. The models used were trained, validated and tested on the same language. \\
We utilized the following language models for our experiments:
\begin{enumerate}
    \item Hate-speech-CNERG/dehatebert-mono-english \footnote{\url{https://huggingface.co/Hate-speech-CNERG/dehatebert-mono-english}}
    \item l3cube-pune/hate-multi-roberta-hasoc-hindi\footnote{\url{l3cube-pune/hate-roberta-hasoc-hindi}}
    \item Hate-speech-CNERG/dehatebert-mono-french\footnote{\url{https://huggingface.co/Hate-speech-CNERG/dehatebert-mono-french}}
    \item Hate-speech-CNERG/dehatebert-mono-german\footnote{\url{https://huggingface.co/Hate-speech-CNERG/dehatebert-mono-german}}
    \item Hate-speech-CNERG/dehatebert-mono-spanish\footnote{\url{https://huggingface.co/Hate-speech-CNERG/dehatebert-mono-spanish}}
\end{enumerate}

Binary hate labels distribution results from monolingual experiments are presented in table \ref{preds_table}

\subsubsection{Machine Translation Experiments}
We adopted the methodology to utilize open source machine translation models for translation of English data to multilingual data and vice-versa. To select the best machine translation model for each of our languages, we evaluated a number of models on a small manually annotated dataset. The translations were carried out for each of the these languages: Hindi, Arabic, French, German and Spanish. Two types of translations were carried out using the translation models: 
\begin{enumerate}
    \item Translation of English dataset collected from various open source hate speech datasets. All the English language samples in this dataset were translated to other 5 languages.
    \item Translation of multilingual data collected from Twitter. This data was collected for the above 5 languages, and each of this language data was translated into corresponding English language data.
\end{enumerate}
Consideration for choosing and evaluating the translation models was based on whether they were open source, free/easy to use, and the translation quality. Translation models used were:
\begin{enumerate}
    \item Google sheets translation 
    \item m2m-100-1.2B\footnote{\url{https://github.com/UKPLab/EasyNMT}} 
    \item IndicTrans\footnote{\url{https://github.com/AI4Bharat/indicTrans}}
\end{enumerate}

The evaluation metrics used were bleu, rougeL and semantic similarity. For semantic similarity, \textit{all-mpnet-base-v2} model from \textit{sentence-transformers} library was used to calculate the cosine score between input and translated sentence embeddings. Figure \ref{translated} shows comparisons of bleu, rougeL and semantic similarity scores for the 3 models on different languages. After comparing the performances and taking other considerations into account, \textit{indicTrans} model was selected for all translations of Indic languages (Hindi), and \textit{m2m-100-1.2B} was selected for the other languages. \\

\pagebreak
\newpage

\begin{figure}[!tbp]
  \centering
  \begin{minipage}[b]{0.5\textwidth}
    \includegraphics[width=\textwidth]{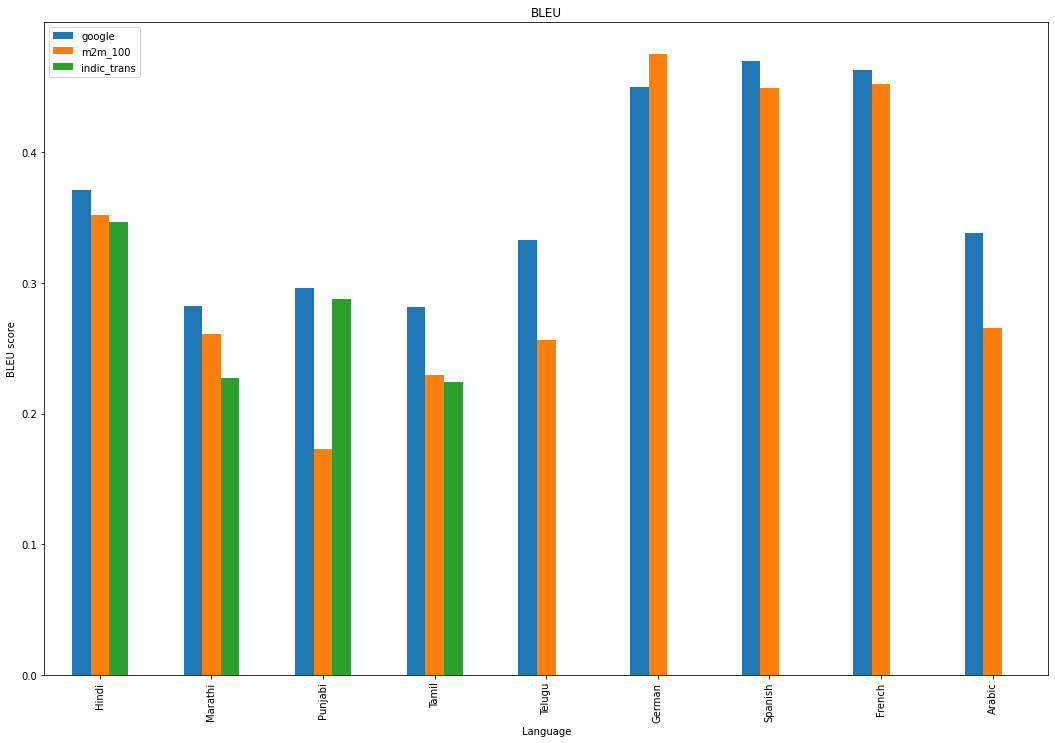}
    \caption{bleu scores for model translations}
  \end{minipage}
  \hfill
  \begin{minipage}[b]{0.5\textwidth}
    \includegraphics[width=\textwidth]{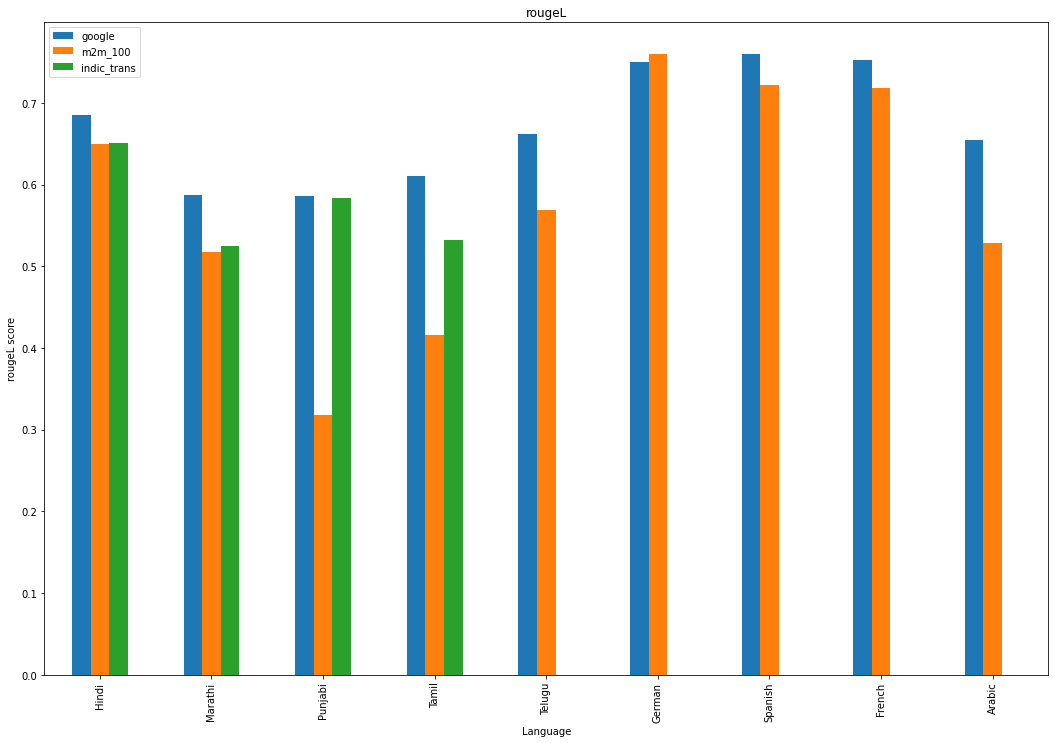}
    \caption{rougeL scores for model translations}
  \end{minipage}
  \hfill
  \begin{minipage}[b]{0.5\textwidth}
    \includegraphics[width=\textwidth]{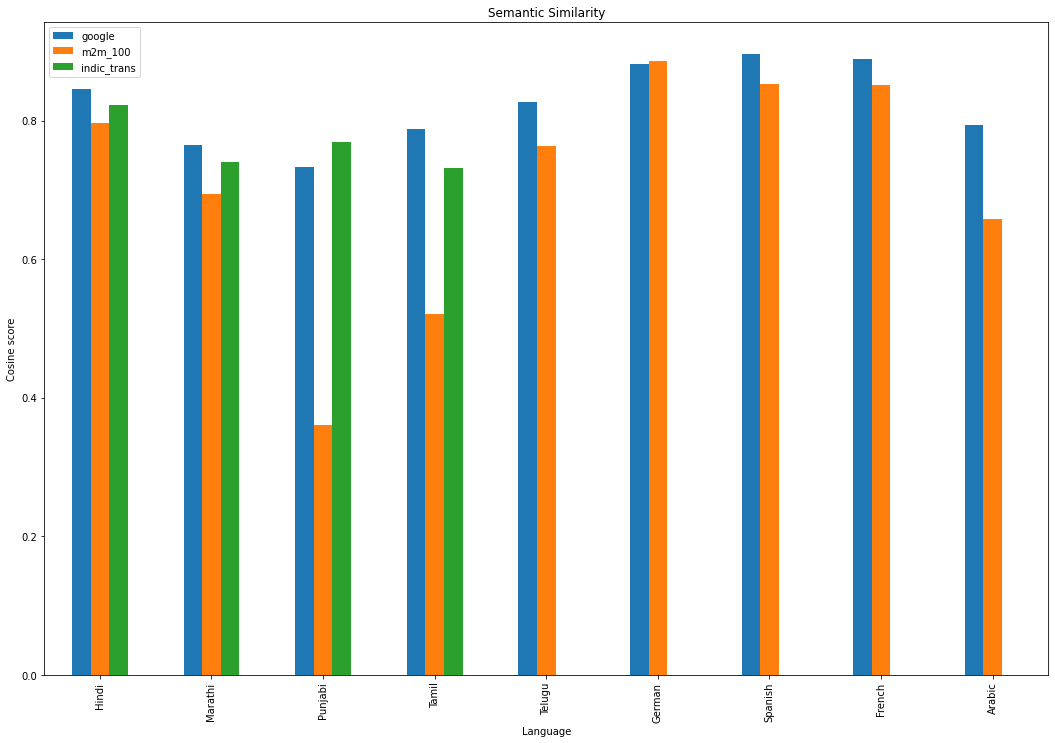}
    \caption{Semantic similarity scores for model translations}
  \end{minipage}
\end{figure}
\label{translated}
\pagebreak
\newpage

\subsubsection{Binary Multilingual Experiments}
We utilized \textit{mbert} which is a pre-trained model on the top 104 languages with the largest Wikipedia corpus using a masked language modeling (MLM) objective, to fine-tune for binary classification task to perform few shot learning experiments. The aim was to make the classifier learn over samples from a few languages and test over other languages. The training dataset was predominant in English, curated from multiple open source resources. In addition to that we introduced language samples for a few other languages using a neural machine translation based open source model ($m2m\_100\_1.2B$)\footnote{\url{https://github.com/pytorch/fairseq/tree/main/examples/m2m_100}}. NMT is an approach to achieve machine translation using artificial neural nets to predict the best possible sequence of words.

The model was trained, tested and validated on the custom multilingual dataset curated with the class distribution shown in Table \ref{Tab mul} per chosen language. We utilized this model to make predictions on the curated dataset.

\begin{table}[h!]
\caption{Multilingual binary classifier training data}
\label{Tab mul}
\begin{tabular}{@{}llll@{}}
\toprule
Languages & Hate & NoHate  & Total \\
\cmidrule(r){1-4}
English   &  20620     & 18171      &  38791     \\
Arabic    &  20539     & 18138       &  38680    \\
Hindi     &  20541     & 18138       &   38679  \\
French    & 20542      &  18138       &  38680  \\ \cmidrule(r){1-4}
\textbf{Total}    & \textbf{82242} & \textbf{72584} &  \textbf{154826} \\ \bottomrule
\end{tabular}
\end{table}

\subsubsection{Perspective API Experiments}

Perspective API is a free to use API that uses machine learning to identify toxicity in the comments. It has API rate limitations of 1 request per second. Due to this limitation we utilized it on all languages except English.
We utilized Google Cloud to generate the Perspective API key for sending requests, and specifically \textit{CommentAnalyzer} API to get the toxicity score for each data point.  
The \textit{CommentAnalyzer} API supports all the 5 languages we have considered. We used the Twitter cleaned data we had collected for the 5 languages to get the Toxicity score to classify each tweet as hate or not-hate.
Numbers are reported in Table \ref{preds_table}.

\subsection{ Multi-Domain Experiments}
This section describes the multi-domain experiments to validate labels on the LAHM dataset.

\subsubsection{Machine Translation Based Multi-Domain Experiments}
HSMerge data present in Table \ref{tab hsmerge} was used for the multi-domain classification of hate speech. This dataset was pre-processed by removing URLs, mentions/usernames, null values and duplicates.

The aim was to utilize transfer learning to fine-tune a model on only the English dataset, and use it to do annotations for the English translations of the data collected from Twitter in other 5 languages.

The model used for fine-tuning on this dataset was\textit{ "distilbert-base-uncased"}\footnote{\url{https://huggingface.co/distilbert-base-uncased}}. DistilBERT is a distilled version of the BERT base model. It has 40 percent less parameters than bert-base-uncased, is 60 percent faster and retains almost 97 percent of BERT’s performances\footnote{\url{https://arxiv.org/abs/1910.01108}}. A linear layer on top of the pooled output of the model was added for the multi-class classification. During training, the maximum sequence length was limited to 256, batch size was set to 32, learning rate 5e-5 and was trained for 4 epochs.
The dataset was split into train, validation and test set using stratified random sampling. Dataset details : 
\begin{enumerate}
\item 20,174 training samples 
\item 5,044 validation samples 
\item 6,305 test samples 
\end{enumerate}

The model evaluation metrics on the test set are given in Table \ref{translated_multiclass met}.

\begin{table}[hp]
\centering
\caption{Translated multiclass metrics}
\label{translated_multiclass met}
\captionsetup{justification=centering}
\begin{tabular}{@{}lllll@{}}
\toprule
Domain & precision & recall & f1-score & support \\ \cmidrule(r){1-5}

racism                & 0.94      & 0.95   & 0.94     & 2882    \\
sexism                & 0.93      & 0.93   & 0.93     & 2295    \\
religious hate              & 0.92      & 0.93   & 0.92     & 409     \\
abuse               & 0.94      & 0.91   & 0.93     & 589     \\
extremism             & 0.83      & 0.82   & 0.82     & 130      \\\cmidrule(r){1-5}

accuracy              &           &        & 0.93     & 6305    \\
macro avg             & 0.91      & 0.91   & 0.91     & 6305    \\
weighted avg          & 0.93      & 0.93   & 0.93     & 6305   \\\bottomrule
\end{tabular}

\end{table}

\subsubsection{Cross lingual based Multi-Domain Experiments }
To validate our fine-grained labels with keywords from hatebase, we experimented with cross-lingual models trained on HSmerge dataset as shown in Table \ref{tab hsmerge}. In this experiment setting we fine tuned the \textit{mbert} model for the multi-class classification task and trained the model on HSmerge data in Table \ref{tab hsmerge} and do the prediction on the other languages. Empirical results can be found under Table \ref{cross_ling res}. In this zero-shot setting, no other language samples were given to the model after fine-tuning on the English dataset and also machine translation was not involved at all in either the pre-training or fine-tuning.

We couldn't manage to maintain the class distribution, and the \textit{extremism} class was left with comparatively less data points. To counter the issue we introduced custom class weights to provide equal attention to the minority class which in our case was \textit{extremism}. This was done using a weighted class random sampler \textit{WeightedSampler} for class imbalance, so that all classes have equal probability. The empirical scores achieved by the cross-lingual model can be found in the Table \ref{cross_ling res}.

The open source {\textit{mbert}} model we fine-tuned for our use case has previously been tested for zero shot experimentation\footnote{\url{https://github.com/google-research/bert/blob/master/multilingual.md##results}} and did manage to achieve decent empirical scores shown in Table \ref{bert_zero_shot}. 
Dataset details :
\begin{enumerate}
\item 19298 training samples 
\item 6432 validation samples 
\item 6434 test samples 
\end{enumerate}

\begin{table}[h!]
    \centering
    \caption{Bert zero shot gold \textbf{F1} scores}
    \label{bert_zero_shot}
    \begin{tabular}{@{}lllll@{}}
        \toprule
        Model & English  & Spanish & German & Arabic  \\ \cmidrule(r){1-5}
        BERT-Zero Shot  & 81.4 & 74.3 & 70.5 & 62.1
         \\\bottomrule
    \end{tabular}

\end{table}

\begin{table}[h!]
 \centering
\caption{HSMerge cross-lingual metrics}
\label{cross_ling res}
\begin{tabular}{@{}lllll@{}}
\toprule
Domains & precision & recall & f1-score & support \\ \cmidrule(r){1-5}

racism                & 0.93      & 0.95   & 0.94     & 2161    \\
sexism                & 0.93      & 0.91   & 0.92     & 1722    \\
religious hate             & 0.93      & 0.93   & 0.93     & 307     \\
abuse              & 0.93      & 0.92   & 0.92     & 442     \\
extremism             & 0.85      & 0.80   & 0.83     & 97      \\\cmidrule(r){1-5}

accuracy              &           &        & 0.93     & 4729    \\
macro avg             & 0.91      & 0.90   & 0.91     & 4729    \\
weighted avg          & 0.93      & 0.93   & 0.93     & 4729   \\\bottomrule
\end{tabular}

\end{table}


\begin{table*}[h!]
\caption{ All models prediction on high level label Hate and NoHate}
\resizebox{\textwidth}{!}{%
\begin{tabular}{@{}lllllllll@{}}
\toprule
\multirow{2}{*}{Language} & \multicolumn{2}{l}{Monolingual} & \multicolumn{2}{l}{Multilingual} & \multicolumn{2}{l}{Perspective} \\ \cmidrule(l){2-7} 
                           & Hate          & NoHate          & Hate           & NoHate          & Hate          & NoHate         &  \\ \cmidrule(r){1-7}
English                    &   7247         &     101345            &    11380            &      99598         &     -          &      -           \\
Hindi                      &    210           &   260              &    3970           &   7247     &   1674              &     3720            \\
Arabic                     &    6469           &   26265              &     3914           &    28820             &    1376           &     31358            \\
French                     &   9148            &     24185            & 8893               &  24440     &    7540           &     13269            \\
German                     &   468            &  23910               &   3905             & 20473        &    4467           &     4164            \\
Spanish                    & 2088              & 10087                &    4734            &   7441          &     24952          &     30196            \\ \bottomrule
\textbf{Total} & \textbf{25630} & \textbf{186052} &  \textbf{36796} & \textbf{188019}&  \textbf{40009} & \textbf{82707} \\

\bottomrule
\end{tabular}%
}
\label{preds_table}
\end{table*}

\begin{table}[]
\caption{Voting across multiple models for hate detection}
\begin{tabular}{@{}llll@{}}
\toprule
Language & Mono-Lingual              & Multi-lingual & Perspective API \\ \midrule
English   & \checkmark &               &  \checkmark               \\
Hindi     &                           &     \checkmark          & \checkmark               \\
Arabic    & \checkmark                         & \checkmark             &                 \\
French    & \checkmark                         & \checkmark             &                 \\
German    &                           & \checkmark            & \checkmark               \\
Spanish   & \checkmark                         &               & \checkmark               \\ \bottomrule
\end{tabular}
\label{tab:voting table}
\end{table}
\hspace{2em}
\begin{table*}[h!]
\caption{Cross-lingual multi-domain predictions distribution}
\resizebox{\textwidth}{!}{%
\begin{tabular}{@{}llllll@{}}
\toprule
Language & Abuse & Sexism & Racism & Religious & Extremism \\
\cmidrule(r){1-6}
English   &  4228     & 2934      &  1566      &   751        &   154        \\
Arabic    &  1895     & 994       &  736      &   164        &    5631       \\
Hindi     &  1511     & 440       &   282     & 22           & 1715           \\
French    & 813      &  779       &  2393   &  750          & 4413           \\
German    &  180     &  123       &  30       &   4        &  131         \\
Spanish   & 818      &  483      &   139     &  11         &   637       \\ \cmidrule(r){1-6}
\label{cross_lingual}
\textbf{Total}    & \textbf{9445} & \textbf{5753} & \textbf{5146} & \textbf{1702} & \textbf{12681} \\ \bottomrule
\end{tabular}%
}
\end{table*}

\begin{table*}[h!]
\centering
\caption{Machine Translation multi-domain predictions distribution}
\resizebox{\textwidth}{!}{%
\begin{tabular}{@{}llllll@{}}
\toprule
Language & Abuse & Sexism & Racism & Religious & Extremism \\
\cmidrule(r){1-6}
English   & 8691      & 6693       & 2495       & 261          & 524          \\
Arabic    & 430      & 1302       & 209       & 259          & 1138         \\
Hindi     & 132      & 204       & 67       & 137          & 105          \\
French    & 745      & 1004       & 545       & 273          & 538 \\
German    & 232      & 608       & 186       & 58          & 98          \\
Spanish   & 2082      & 2024       & 1519       & 347          & 1310
   \\ \cmidrule(r){1-6}

\textbf{Total}    & \textbf{12312} & \textbf{11835} & \textbf{5021} & \textbf{1335} & \textbf{3713} \\ \bottomrule
\end{tabular}%
}
\label{multi_class}
\end{table*}


\section {Evaluation}

We performed initial level of validation (Hatefulness) experiments on the raw LAHM dataset using open source monolingual models, Perspective API and the multilingual binary classifier we trained. We compared the results for monolingual and multilingual and noticed that our multilingual few-shot learning based classifier out-performed the open source monolingual BERT based models from HuggingFace except on Arabic and French languages where the latter did a better job; details can be found in Table \ref{preds_table}. Based on the analysis we took at least 2 votes from the models which performed better, details shown in Table \ref{tab:voting table}.

For Hindi, Perspective API performed better by predicting 31 percent as hate compared to 12 percent for binary classifier. For Arabic, the binary classifier performed significantly better.

For domain validation we utilised the zero-shot cross-lingual,  multi-domain and machine translation based multi-domain models trained with HSmerge data in Table \ref{tab hsmerge} and performed predictions on the different languages.
Label distribution from cross-lingual models are shown in Table \ref{cross_lingual}.
Label distribution of predictions for different languages on the LAHM dataset from multi-class translated model is in Table \ref{multi_class}.

For Hindi, 31.6 percent samples were Abuse, while the lowest were 10.3 percent for Racism.
For Arabic, Sexism and Extremism contributed to 73 percent of all hate samples.
For French, 32 percent were Abuse and lowest 8.3 percent were Religious Hate.
For German, 51 percent were Sexism while the lowest were Religious Hate and Extremism together amounting to 13 percent.
For Spanish, Abuse and Sexism had 27 percent each, while Religious Hate had 5 percent of samples.

\section {Conclusion}
We have presented the LAHM dataset, a large scale semi-supervised training dataset for multilingual and multi-domain hate speech identification, we created by using 3 layer annotation pipeline and combination of monolingual, multilingual and cross-lingual models. To the best of our knowledge, LAHM is the largest of its kind, containing close to 300k tweets across 6 languages and 5 domains.

LAHM enables cross-lingual abusive language detection across five domains and in-depth interplay between language shift and domain shift. We have profiled LAHM as a comprehensive resource for evaluating hate speech detection through a series of cross-domain experiments in monolingual, multilingual and cross-lingual setups with state of the art transfer learning models.

We hope that LAHM will inspire more efforts in understanding and building semi-supervised large scale multilingual and multi-domain abusive language detection datasets.\\

\section {Future Work}

For future work we explored leetspeak detection and identification on social networks. A lot of hate content on social media uses leetspeak to evade moderators and automated systems. We collected hate keywords belonging in this category for each of the 5 domains, and experimented with a set of leets for each keyword to extract the leetspeak hate content. We plan to use this for future work in multilingual and multi-domain settings.

\bibliography{LAHM}

\appendix

\end{document}